# Multimodal MRI brain tumor segmentation using random forests with features learned from fully convolutional neural network

Mohammadreza Soltaninejad, Lei Zhang, Tryphon Lambrou, Nigel Allinson, Xujiong Ye

Laboratory of Vision Engineering, School of Computer Science, University of Lincoln, UK

**Abstract.** In this paper, we propose a novel learning based method for automated segmentation of brain tumor in multimodal MRI images. The machine learned features from fully convolutional neural network (FCN) and hand-designed texton features are used to classify the MRI image voxels. The score map with pixel-wise predictions is used as a feature map which is learned from multimodal MRI training dataset using the FCN. The learned features are then applied to random forests to classify each MRI image voxel into normal brain tissues and different parts of tumor. The method was evaluated on BRATS 2013 challenge dataset. The results show that the application of the random forest classifier to multimodal MRI images using machine-learned features based on FCN and hand-designed features based on textons provides promising segmentations. The Dice overlap measure for automatic brain tumor segmentation against ground truth is 0.88, 080 and 0.73 for complete tumor, core and enhancing tumor, respectively.

## 1   Introduction

Segmentation of brain tumors from multimodal magnetic resonance imaging (MRI) is a challenging task due to different types and their complicated structures in the images [1] and also large variety and complexity within one type of tumor in terms of characteristics such as intensity, texture, shape and location. The challenge is developing a platform which creates accurate segmentation and works for multiple tumor types and different imaging equipment [2].

In recent decades, the research work for automatic brain tumor segmentation has increased which represents the demand for this area of research and it is still in progress [3]. Several methods have been proposed in the literature for detection and segmentation of tumors in MRI images [4]. The brain tumor segmentation techniques can be categorized into generative and discriminative based models [3].

Discriminative approaches are based on extraction of the features from the images and creating models based on the relationship between the image features and the voxel classes. A vast variety of features are used in the literature such as intensity based [5], histogram based [6] and texture features [7]. Most of the brain tumor segmentation techniques used hand designed features which are fed into a classifier such as random forests (RF) [8, 9]. Among the conventional classifiers, RFs presents the best segmentation results [3, 9]. A limitation of discriminative approaches which use hand designed feature is that in order to offer better description of the tissues in the images, they are required to use a large number of features which results in high dimensional problems which make the process more complicated and time consuming. In addition, a large number of experiments and optimization should be conducted in order to identify the optimum parameters for feature extraction and also the optimum classifier.

To tackle this problem, another variant of discriminative approaches which is using convolutional neural networks (CNN) for medical image analysis has attracted significant attention in recent years. Several methods have developed CNNs to segment the brain tumors in MRI more accurately [10, 11]. A limitation of CNN based methods is that the classification is performed on the voxels, therefore the local dependencies are not taken into account. Whilst some hand designed feature extraction methods consider the spatial features and local dependencies of the voxel classes.

In this paper, we proposed a novel learning based method for fully automated segmentation of brain tumor in multimodal MRI images. The machine-learned features from fully convolutional neural network (FCN) were used to classify the MRI image voxels. The score map extracted from the FCN were used to localize the tumor area and also as a feature extraction section too for the further classification stage. To consider the voxel neighborhood system and also more accurate classification, texton based features were used. The proposed method was applied on the publicly available BRATS 2013 dataset [12, 13]. The segmentation results presented here were provided by the online system. It should be noted that the ground-truth for the challenge dataset are not available, and that the evaluation was performed by uploading our segmentation and getting the evaluation results from the online system.

The main contributions of our method can be summarized as follows:

- Proposing a novel fully automatic learning based segmentation method, by applying the machine-learned features to the state-of the art random forest classifier.
- Applying hand designed texton based features while considering the spatial features and local dependencies in order to improve the segmentation accuracy.
- Using the FCN to include the tumor region and locally focusing on more accurate segmentation of tumor in a reasonable processing time by excluding unnecessary processing of other parts of the brain that are detected as normal by FCN.

## 2 Method

Our method is comprised of four major steps (pre-processing, FCN, Texton map generation, and RF classification) that are depicted in Fig. 1.

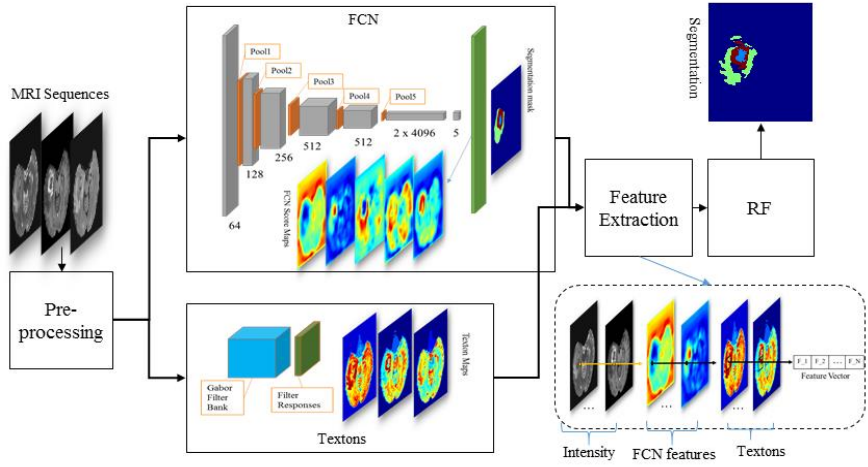

**Fig. 1.** Flowchart of the proposed method. The FCN architecture and feature extraction procedure.

### 2.1 Preprocessing

In the first instance we exclude the 1% highest and lowest intensity values for each image. The intensities are then normalized for each protocol by subtracting the average of intensities of the image and dividing by their standard deviation. In turn, for each individual protocol, the histogram of each image is matched to the one of the patient images which is selected as the reference and then the dynamic range of the intensities is linearly normalized to the range [0, 1].

### 2.2 Fully Convolutional Neural Network

In this paper, we adopted FCN-8s architecture in [14] for segmentation of brain tumor in multimodal MRI images, where the VGG16 [15] is employed as CNN classification net. In the FCN architecture, initially, the classification net is transformed to be fully convolutional net, then adding an upsampling or de-convolutional layer to it for pixel wise predictions. The FCN training is end-to-end supervised learning procedure and the image segmentation is performed using a voxel-wise prediction /classification. The FCN-8s constructed from FCN-16s skip net and FCN-32s coarse net. The predictions at shallow layers are produced using skip layer which combines coarse predictions at deep layers to improve segmentation details. More specifically in our experiment, the FCN-8s is implemented by fusing predictions of shallower layer (Pool3) with 2 × upsampling of the sum of two predictions derived from pool4 and last layer. Then the stride 8 predictions are upsampled back to the image.

The FCN-8s produces more detailed segmentations comparing to the FCN-16s, however, the lack of the spatial regularization for FCN leads to label disagreement between similar pixels and diminished the spatial consistency for segmentation.

In the next section, we introduce spatial features extracted based on texton which uses the three dimensional connectivity neighborhood system to complement the FCN weak point of considering only the voxels.

### 2.3 Spatial texton features

The texton based features are strong tools for description of textures in images. They are applied to the proposed method as human-designed features to support the machine-learned features and improve the segmentation results. Textons are obtained by convolving the image with a specific filter bank. In this paper, Gabor filters are used which are defined by the following formulation:

$$G(x,y;\theta,\sigma,\lambda,\psi,\gamma) = \exp(-\frac{x'^2+\gamma^2 y'^2}{2\sigma^2})\exp(i(2\pi\frac{x'}{\lambda}+\psi)) \quad (1)$$

where, σ is the standard deviation of Gaussian envelope, γ is the spatial aspect ratio, λ is the wavelength of sinusoid and ψ is the phase shift. The terms x' and y' are obtained from:

$$\begin{cases} x' = x\cos\theta + y\sin\theta \\ y' = -x\sin\theta + y\cos\theta \end{cases} \quad (2)$$

where, θ is the spatial orientation of the filter. A set of Gabor filters with different parameters creates the filter bank. The Gabor filter parameters are chosen using exhaustive grid search. To cover all orientations six different filter directions were used: [0º, 30º, 45º, 60º, 90º, 120º]. Filter sizes are in the range from 0.3 to 1.5 using a step of 0.3. The wavelength of sinusoid coefficients of the Gabor filters were 0.8, 1.0, 1.2 and 1.5.

Each MRI protocol is convolved with all the Gabor filters in the bank. The filter responses are then merged together and clustered into $k_{texton}$ clusters using *k-means* clustering. The number $k_{texton}$ = 16 was selected as the optimum value for the number of clusters in texton map. The texton map is created by assigning the cluster number to each voxel of the image. The texton feature for each voxel is the histogram of textons in a neighborhood window of 5 × 5 around that voxel.

The feature vector is generated for each voxel based on the score map from the FCN. For each class label, a score map is generated, 5 maps are generated using the standard BRATS labelling system. The values of each map layer corresponding to each voxel are considered as machine-designed features of that voxel. The normalized intensity value of the voxels in each modality which is obtained from the pre-processing stage is also included in the feature vector. Therefore, in total 56 features were collected (5 FCN score map, 3 protocol intensity and 48 texton histogram) for usage in the next step.

### 2.4 RF classification

Random forests (RF) is among the most powerful classifiers [16], and is an ensemble of multiple decision trees. Each node of a tree includes a set of training examples and a predictor. A random subset of features is selected at each attribute split during the bagging process. The trees are growing until a specified tree depth $D_{tree}$. A vote for the most popular class is made after generating a large number of trees. The target region in which the RF is applied is guided by the ROI which was detected by the FCN. A confidence margin of 10 voxels in 3D space around the detected tumor area is selected by morphological dilation. The feature vector for voxels in this target area are fed to the random forests for training. The main parameters in designing RF are the number of trees, tree depth and the number of attributes ($k_{atribute}$) which is selected to perform the random split. The optimum value for $k_{atribute}$ for the classification tasks is $k_{atribute} = \sqrt{N_{feature}}$ where $N_{feature}$ is the total number of features, in our study $k_{atribute} = 7$. RF parameters were tuned by examining different tree depths and number of trees on clinical training datasets and evaluating the classification accuracy using 4-fold cross validation. The number of trees $N_{tree} = 50$ with depth $D_{tree} = 15$ provide an optimum generalization and accuracy. Based on the classes assigned for each voxel in the test dataset, the final segmentation mask is created by mapping back the voxel estimated class to the segmentation mask volume. Finally, the bright regions in the healthy part of the brain near to the skull are eliminated using a connected component analysis.

## 3  Experimental Results and Discussion

The method was evaluated on the publicly available MICCAI BRATS 2013 [12, 13] dataset which is provided by Virtual Skeleton Database (VSD) [13]. The training dataset consists of 30 patient MRI scans of which 20 are high-grade and 10 are low-grade gliomas. The test dataset consists of 10 cases with high grade gliomas. The dataset has been already skull-removed, registered and interpolated by the BRATS challenge organizers. The MRI sequences FLAIR, T1-weighted+contrast and T2-weighted are applied to the FCN. The segmented masks obtained from our automated method using the challenge testing dataset are uploaded to the website and evaluated by the corresponding online system. The ground-truth for training datasets are provided in which four labels are assigned to the tumor tissue parts i.e. oedema, necrosis, enhancing and non-enhancing tumor. In our method we use the BRATS challenge standard combination which are enhancing tumor, core (including necrosis, enhancing and non-enhancing) and complete tumor.

The proposed method was performed on MATLAB 2016b on a PC with CPU Intel Core i7 and RAM 16 GB with the operating system windows 8.1. The FCN was implemented using MatCovNet toolbox [17]. GPU GeForce gtx980i was used for reducing the training time of the FCN. The RF was implemented using open source code provided in [18] which is a specialized toolbox for RF classification based on MATLAB.

The evaluation measure which are provided by the VSD website, i.e. Dice score, positive predictive value (PPV) and sensitivity were used to compare the segmentation results with the gold standard (blind testing). Table 1 provides the evaluation results obtained by applying the proposed method on BRATS 2013 challenge dataset. In the third row of Table 1, the values in parentheses show the current ranking of each individual measure for the corresponding tumor part in the VSD website at the time of submission. Currently our overall rank is 5[th] for the challenge dataset.

Fig. 2 shows examples of segmentation of tumors parts in some BRATS 2013 challenge dataset using FCN only and our proposed method. As the ground truth is not accessible we are not able to include it in Fig. 2.

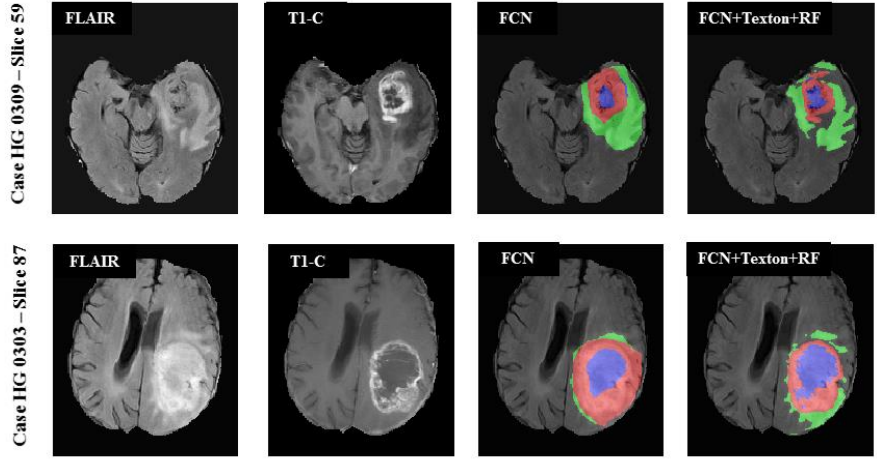

**Fig. 2.** Segmentation results for some cases of BRATS 2013 challenge dataset. The first column shows the original FLAIR images, the second shows T1-weighted-contrast, the third column shows the segmentation mask of FCN overlaid on FLAIR image and the forth column shows the segmentation mask of the proposed method overlaid on FLAIR image. Oedema: green, necrosis: blue, enhancing tumor: red and on-enhancing tumor: yellow.

In order to evaluate the performance of the proposed method (FCN+Texton+RF), two comparative experiments were also set up. In the first scenario, the labels which were directly classified by FCN were considered as the final segmentation mask. In the second scenario, the images were segmented with the features from the FCN score maps and then classified by RF (FCN+RF). Table 1 shows our final experimental results.

It can be seen that the sensitivity is very good for segmentation using FCN only but the Dice overlap and PPV are not so good. This implies that FCN is able to detect the area which includes the tumor, but it is not able to accurately and locally detect its boundaries. It can be seen in the third column of Fig. 2 that the FCN over-segmented the tumor area especially the tumor core. Using the machine learned features and application of RF there is a slight improvement of the Dice score for complete tumor and significantly improvement for tumor core and enhancing part and also improvement of PPV for all areas. It means that the segmentation boundaries are now closer to the ground truth. Sensitivity for complete tumor decreases which represent under-segmentation. Adding the texton features to the pipeline improves the overlap measure for complete tumor and increases the sensitivity while slightly decreases the PPV. Therefore, the proposed method improves the overlap measure while maintaining a balance between sensitivity and PPV.

The FCN segmentation was able to locate the tumor areas, but is not able to accurately segment them, as it provides coarse segmentations (see Fig.2. third column). To tackle this problem, we propose to consider the local dependencies and neighborhood system of voxels in classification by using texton features. The experimental results emphasises that this refinement increases significantly the accuracy while keeping balance between sensitivity and PPV. As an example we can observe that our FCN+Texton+RF method produces finer segmentations compared with the segmentations of FCN+RF. Using 3D texton also enable us to consider the connectivity information in all directions in 3D space which compensate the limitation of FCN which only works on 2D slices.

One limitation of the proposed method is that the training stage is time consuming. However, when the model is created, both FCN and RF are fast to use for classification of new datasets. Also the model can be saved so any future training dataset, can be added to the previously trained model and update it.

The results of our proposed method which is applied on BRATS 2013 clinical dataset and the related top-ranked works on the same dataset which are on the website scoreboard [13] are presented in Table 1. The method in [11] used a developed version of deep convolutional neural network. The method proposed by Pereira [10], which is based on CNN has the best score and ranking on the VSD scoreboard. Our proposed method has the same Dice score, PPV and sensitivity for the complete tumor to this method. The method proposed by Tustison et al. [19], which used RF and hand designed features, was the winner of the on-site BRATS 2013 challenge. Our method outperformed [19] in terms of Dice score for complete tumor and core and PPV value for all tumor tissue types. Our method has the best dice score which is 0.88 at the time of this submission.

**Table 1.** Segmentation results per case for BRATS 2013 challenge dataset which is evaluated by VSD website. Comparison with other works which used BRATS 2013 challenge dataset and are top ranked.

| Method | Dice score | | | Positive Predictive Value | | | Sensitivity | | |
|---|---|---|---|---|---|---|---|---|---|
| | Complete | Core | Enhancing | Complete | Core | Enhancing | Complete | Core | Enhancing |
| FCN | 0.79 | 0.69 | 0.62 | 0.77 | 0.68 | 0.58 | 0.83 | 0.82 | 0.75 |
| FCN + RF | 0.80 | 0.83 | 0.71 | 0.90 | 0.90 | 0.73 | 0.74 | 0.78 | 0.78 |
| **FCN+Texton + RF** | **0.88 (1)** | **0.80 (9)** | **0.73 (19)** | **0.88 (11)** | **0.87 (8)** | **0.80 (5)** | **0.89 (21)** | **0.77 (21)** | **0.70 (32)** |
| Havaei [11] | 0.88 | 0.79 | 0.73 | 0.89 | 0.79 | 0.68 | 0.87 | 0.79 | 0.88 |
| Tustison [19] | 0.87 | 0.78 | 0.74 | 0.85 | 0.74 | 0.69 | 0.89 | 0.88 | 0.87 |
| Pereira [10] | 0.88 | 0.83 | 0.77 | 0.88 | 0.87 | 0.74 | 0.89 | 0.83 | 0.81 |

# 4   Conclusion

In this paper, a learning based automatic method is proposed for segmentation of brain tumor in MRI images. The method is a hybrid approach in which the machine-learned features extracted using the FCN are used alongside with hand designed texton features and applied to the state-of-the-art RF classifier. The proposed method was evaluated on BRATS 2013 challenge dataset by the provided online system. The experimental results suggest that the proposed method achieves promising results in the segmentation of brain tumor and its parts. Adding texton features from different protocols to the system increases the classification accuracy of the voxels and the final segmentations.